\documentclass{article}
\usepackage{spconf,amsmath,graphicx,hyperref}
\usepackage{booktabs}

\title{CASA: Content-Acoustic Speaking Assessment with Speech Encoder and Large Language Model}

\name{
\begin{tabular}{c}
Nhan Phan$^{1}$ \qquad
Ilona Lähteenmäki$^{2}$ \qquad
Anna von Zansen$^{2}$ \qquad
Olli-Pekka Pauna$^{1}$ \\
Yaroslav Getman$^{1}$ \qquad
Tamás Grósz$^{1,3}$ \qquad
Mikko Kurimo$^{1}$
\end{tabular}
}

\address{
$^{1}$ Department of Information and Communications Engineering,
Aalto University, Espoo, Finland \\
$^{2}$ Department of Education,
University of Helsinki, Helsinki, Finland \\
$^{3}$ Programmable Autonomous Systems Division,
Walton Institute, Waterford, Ireland
}

\begin{document}
%
\maketitle
%

\begin{abstract}

Research on automatic speaking assessment (ASA) has increasingly adopted multimodal speech large language models to assess learners' speaking performance. However, existing studies provide limited analysis of how acoustic and content information contribute to predictions and how stable the resulting performance is. We propose CASA, a simpler architecture combining Whisper-medium and Qwen3.5-2B that achieves state-of-the-art performance while providing a more interpretable separation between speech delivery and content.

On the Speak \& Improve Corpus 2025, CASA achieves a root mean square error (RMSE) of 0.358, improving on the previous best RMSE while using approximately half the estimated inference parameters. The general-purpose architecture is designed for adaptation to other ASA corpora without structural changes and relies on three handcrafted fluency features. Through ablations and repeated runs, we analyze the individual and complementary contributions of acoustic and content information, examine performance variability, and demonstrate the potential of large language model reasoning for training-free content validation.
\end{abstract}
\section{Introduction}
\label{sec:intro}

Automatic speaking assessment (ASA) aims to automatically estimate second-language (L2) learners' oral proficiency from their spoken responses. Compared with assessments conducted exclusively by human raters, ASA can reduce scoring costs, improve scoring consistency, and enable scalable and timely feedback. Nevertheless, holistic speaking assessment remains challenging because speaking proficiency is reflected in multiple interacting sources of evidence, including speech delivery (how something is said) and speech content (what is said). 

With recent advances in large language models (LLMs), speech LLMs have increasingly been explored for ASA. Ma et al. \cite{maAssessmentL2Oral2025} used Qwen2-Audio for L2 oral proficiency assessment and achieved state-of-the-art (SOTA) performance on the S\&I test set \cite{knill_SpeakImprove2025}, obtaining a Pearson correlation coefficient (PCC) of 0.833; RMSE was not reported. More recently, Lin et al. \cite{linSessionLevelSpokenLanguage2026} combined Phi-4-Multimodal with a Whisper-large encoder \cite{radford_whisper2023}, achieving an RMSE of 0.360 on the same test set. In contrast, Cai et al. \cite{caiTeamPerezososASR2025} achieved a comparable RMSE of 0.364 without using a speech LLM by combining a Whisper-medium encoder and BERT-based \cite{devlin_BERT2019} graders with handcrafted acoustic and linguistic features.

Despite their strong performance, speech-LLM-based graders rely on large multimodal backbones, which increase computational requirements during inference and may limit practical deployment. Although Cai et al. \cite{caiTeamPerezososASR2025} avoid using a speech LLM, their system combines multiple graders and handcrafted features. It also relies on a separate ASR model to produce more verbatim transcripts, increasing pipeline complexity and potentially limiting portability. Moreover, previous studies provide limited analysis of the respective contributions of acoustic and textual information. As their implementations have not been publicly released, reproducibility and further investigation are also limited.

In this work, we propose CASA (content-acoustic speaking assessment), a compact, general-purpose two-branch ASA architecture that explicitly separates acoustic information from linguistic information represented in ASR transcripts. The model employs a Whisper-medium encoder for the acoustic branch, and Qwen3.5-2B \cite{qwen3.5} for the content branch, making it substantially smaller than existing speech-LLM-based systems. Through detailed ablation and component-level analyses, we investigate the individual and complementary contributions of acoustic and textual information. Our model achieves an RMSE of 0.358 on the S\&I test set, slightly outperforming the SOTA result of 0.360 while requiring substantially fewer parameters during inference. The architecture is designed as a general-purpose ASA framework that is not tailored to any dataset and uses only three simple fluency features. Our code is publicly available\footnote{\url{https://github.com/aalto-speech/CASA/}}.

\section{Method}
\label{sec:method}

\subsection{Dataset}
\label{sub:dataset}

The spoken language assessment (SLA) task of the S\&I Corpus \cite{knill_SpeakImprove2025} consists of four parts: Parts 1, 3, 4, and 5. Part 1 contains six short questions, with responses typically lasting 10--20 seconds. Part 3 consists of one-minute open-ended questions in which speakers express their opinions, whereas Part 4 contains one-minute tasks requiring speakers to describe a process illustrated in a graphical prompt. Finally, Part 5 contains five opinion-based questions, each with a maximum response time of 20 seconds. In total, the corpus contains approximately 315 hours of speech and is divided into training, development, and test sets, with the score distributions kept as balanced as possible across the splits. Each part was rated by human assessors and assigned a holistic score based on the CEFR scale \cite{CEFR_2001}. The CEFR levels map to a 2.0–5.5 scale in 0.5 steps (A2=2.0, A2+=2.5, …, C1+=5.5); the final score is the arithmetic mean of the four part scores.

\subsection{Model}
\label{sub:model}

\begin{figure*}[t]
  \centering
  \includegraphics[width=.85\linewidth]{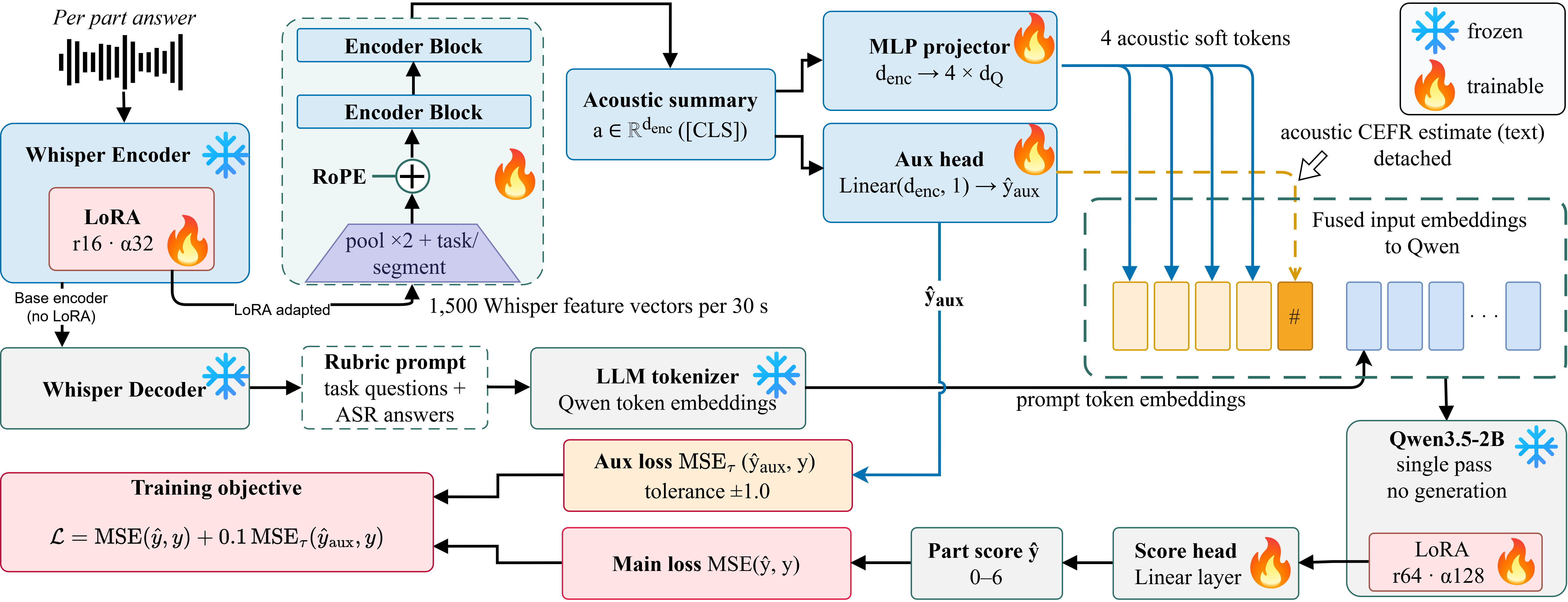}
  \caption{An overview of the model architecture.}
  \label{fig:architecture}
\end{figure*}

Figure~\ref{fig:architecture} presents the proposed model architecture. To support adaptation to other ASA settings, a single shared model scores each part independently. The overall score is then calculated as the arithmetic mean of the per-part predictions.

Our design builds on Phan et al.~\cite{phanOneWhisperGrade2025}, who showed that a single Whisper-small encoder achieves strong performance but still lags behind SOTA~\cite{qian_SpeakImprove2025}. Their model uses global average pooling, which may discard fine-grained acoustic variation. We instead adopt a Whisper-medium encoder, kept frozen: the encoder--decoder still produces the ASR transcript, while LoRA adapters~\cite{hu_lora} on the encoder produce the representation used by the acoustic branch.

\textit{The acoustic branch}, originating from the Whisper encoder, assesses the speaker's delivery. As the S\&I corpus provides no separate delivery score, we supervise it using the holistic per-part CEFR label---the main loss target $y$---which the branch must predict from acoustics alone.

Because Whisper processes at most 30\,s at a time, each answer is split into 30\,s chunks up to a 2\,min cap, with each chunk yielding $1{,}500$ frame-level vectors. Rather than global average pooling, which discards acoustic detail~\cite{phanOneWhisperGrade2025}, we concatenate the chunk encodings and average adjacent frame pairs, reducing the temporal resolution from 20\,ms to 40\,ms and halving the sequence length while preserving most acoustic information.

Parts P1 and P5 comprise several short answers, so before the Transformer we add two zero-initialized, learned embeddings to every frame: a \emph{task} embedding (which part, P1--P5) and a \emph{segment} embedding (which answer within the part). This is intended to let the aggregator distinguish frames from different answers.

The tagged frames pass through a two-layer Transformer encoder~\cite{vaswaniAttentionAllYou2017} with rotary position embeddings (RoPE)~\cite{su_rotary_RoPE2024}, which help preserve positional information over long concatenated sequences. Although this is less critical for the current S\&I data, it supports future adaptation to longer responses. Following the [CLS] pooling of BERT, shown to transfer well to audio representation learning~\cite{liATSTAudioRepresentation2022}, we summarize the sequence with a [CLS] token.

The acoustic summary feeds two heads. An MLP projector maps it to four acoustic soft tokens for the LLM, allowing the acoustic information to be represented across multiple embeddings while remaining compact.  A linear auxiliary head maps it to a scalar CEFR estimate that is detached and rendered as text (e.g., \texttt{acoustic\_\allowbreak cefr\_estimate:\ 4.5}) in the LLM prompt, and also drives the auxiliary loss. Both objectives backpropagate into the shared branch, so the main loss (through the soft tokens) and the auxiliary loss (through the auxiliary head) jointly shape the acoustic representation. Since acoustics alone cannot determine CEFR, we keep the auxiliary term from dominating by assigning it a small weight ($0.1$) and, more importantly, use a tolerance of $\pm 1$ score point. Predictions within this margin incur no auxiliary loss, and only deviations beyond the margin are penalized. Among the tested auxiliary-loss weights and tolerance levels, this combination achieved the best performance.

On \textit{the content branch}, the frozen Whisper encoder--decoder transcribes each answer. The transcripts are generated offline before training for computational efficiency. For each part, the transcript is paired with its task prompt using \texttt{<TASK part=P1>} followed by the corresponding \texttt{Q1/A1} pairs. Qwen3.5-2B receives the fused input sequence in the following order: the four acoustic soft tokens, the detached acoustic CEFR estimate represented as text, the scoring rubric, the question--answer pairs, and three simple fluency statistics derived from the audio and ASR output: duration, silence ratio, and speech rate (words/s). The base Qwen parameters remain frozen and are adapted with another LoRA adapter. The model performs a single forward pass without text generation, and a linear regression head applied to the final-token representation predicts the part score $\hat{y}$.

The training objective combines the main loss from the fused scorer with the  auxiliary loss from the acoustic branch:
\begin{equation*}
  \mathcal{L} = \mathrm{MSE}(\hat{y}, y)
  + 0.1\,\mathrm{MSE}_{\tau}(\hat{y}_{\mathrm{aux}}, y),
\end{equation*}
\begin{equation*}
  \mathrm{MSE}_{\tau}(a,b) = \big(\max(|a-b| - \tau,\, 0)\big)^2,
  \quad \tau = 1,
\end{equation*}
where $\hat{y}$ is the final part prediction, $\hat{y}_{\mathrm{aux}}$ is the acoustic-only prediction, and $y$ is the ground-truth per-part CEFR score.

\section{Results}
\label{sec:results}

Training was conducted on a single NVIDIA H100 80\,GB GPU with a batch size of 16 and gradient accumulation over two steps. Learning rates were $2\times10^{-4}$ (acoustic LoRA), $1\times10^{-4}$ (LLM LoRA), $5\times10^{-5}$ (other modules). Training took approximately two hours. Table~\ref{tab:results} reports the performance of CASA, with the model configuration and checkpoint selected solely according to overall RMSE on the development set. CASA achieves an RMSE of 0.358, marginally lower than the previous SOTA result of 0.360. As this difference is small, we do not claim a meaningful improvement in accuracy. Instead, CASA provides a better accuracy--model-size trade-off, achieving comparable SOTA performance with approximately half the estimated inference parameters of the NTNU speech-LLM system, as shown in Table~\ref{tab:size}. Team Perezoso~\cite{caiTeamPerezososASR2025} can likewise be considered SOTA-level despite not using an LLM, although its pipeline combines multiple graders, handcrafted features, and a separate fine-tuned Parakeet-TDT-1.1B ASR model. One Whisper~\cite{phanOneWhisperGrade2025} remains the most compact system, using only a Whisper-small encoder, although with a higher RMSE of 0.372.

\begin{table}[t]
  \caption{Overall performance on the S\&I test set.
  CASA-Crisper uses CrisperWhisper \cite{zusag_CrisperWhisper_2024}.}
  \label{tab:results}
  \centering
  \begin{tabular}{lcccc}
    \toprule
    Model & RMSE & PCC & \%$\leq$0.5 & \%$\leq$1.0 \\
    \midrule
    NTNU~\cite{linSessionLevelSpokenLanguage2026}
      & 0.360 & 0.827 & \textbf{85.7} & 99.0 \\    
    Perezoso~\cite{caiTeamPerezososASR2025}
      & 0.364 & 0.826 & 83.0 & \textbf{99.7} \\
    \midrule  
    CASA
      & \textbf{0.358} & 0.829 & 84.7 & 98.7 \\
    CASA-Crisper
      & 0.363 & \textbf{0.836} & 84.0 & \textbf{99.7} \\
    \bottomrule
  \end{tabular}
\end{table}

\begin{table}[t]
  \caption{Model size vs.\ performance on the S\&I test set. NTNU and Perezoso
parameter counts are estimated from their described architectures (no released code).}
  \label{tab:size}
  \centering
  \begin{tabular}{lcr}
    \toprule
    Model & RMSE &  Total parameter \\
    \midrule    
    CASA                                    &  0.358 & 3.13\,B \\
    NTNU~\cite{linSessionLevelSpokenLanguage2026} & 0.360 & 6.24\,B \\
    Perezoso~\cite{caiTeamPerezososASR2025} &  0.364 & 2.17\,B \\
    One Whisper~\cite{phanOneWhisperGrade2025} &  0.372 & 0.17\,B \\
    \bottomrule
  \end{tabular}
\end{table}

In terms of architecture, CASA is closer to Cai et al. \cite{caiTeamPerezososASR2025} than to speech-LLM approaches \cite{maAssessmentL2Oral2025,linSessionLevelSpokenLanguage2026}. Cai et al. probe the final six layers of a frozen Whisper-medium encoder, while Lin et al. derive an acoustic proficiency prior from the final representation of a frozen Whisper-large encoder. CASA instead adapts the Whisper-medium encoder across its layers using LoRA. Since acoustic and semantic information is distributed unevenly across Whisper layers \cite{glazer_whisper_layers_2026}, this motivates CASA's task-specific adaptation rather than reliance on a fixed final-layer representation. Cai et al. further combine their Whisper grader with several BERT-based graders and handcrafted features, whereas CASA integrates the acoustic representation with an LLM-based content branch and a tolerance-based auxiliary objective. Apart from three fluency statistics, CASA requires no handcrafted assessment features.

\section{Analysis}
\label{sec:analysis}

The modular acoustic encoder allows us to swap in CrisperWhisper, a verbatim-transcription Whisper-large that preserves the disfluencies that standard Whisper smooths away. As we use the same hyperparameters as for Whisper-medium, the overall RMSE may be worse than what proper tuning would yield. In Table~\ref{tab:perlevel}, CASA-Crisper improves on A2, indicating that verbatim transcripts help the system catch the content errors of A2 and B1 speakers. This gain matters because A2 is an under-represented class. B2 and C1, however, get worse: for otherwise fluent speakers, verbatim transcription may introduce inauthentic errors that hide the range marking the top band. 

We also explore two self-supervised learning (SSL) models: WavLM~\cite{chen_WavLM_2022} and wav2vec2 XLS-R 300M~\cite{babuXLSRSelfsupervisedCrosslingual2022}, swapping only the acoustic encoder while keeping CASA's Whisper ASR transcript, so the acoustic representation is the sole difference from CASA. Both substantially underperform CASA, and an English-finetuned variant \cite{grosman2021xlsr53-large-english} also provides no improvement. We attribute this gap partly to architectural mismatch: CASA's aggregator and training recipe were developed for Whisper representations and may not transfer optimally to SSL encoders. SSL encoders are effective for ASA in other architectures~\cite{banno_ASA_SSL2023}; adapting CASA to exploit them (e.g.\ a dedicated pronunciation head) is left to future work.

Neither a larger acoustic encoder (CrisperWhisper; Table~\ref{tab:results}) nor a larger LLM (Qwen3.5-4B, RMSE 0.364) improves RMSE; even pairing Whisper-large-v3 with the 4B LLM yields only 0.362. Performance thus appears not to be limited by model capacity, though larger models may require different hyperparameters.

\begin{table}[t]
  \caption{Overall RMSE by gold CEFR band on S\&I; Macro is the
  unweighted mean RMSE over each CEFR level.}
  \label{tab:perlevel}
  \centering
  \begin{tabular}{lccccc}
    \toprule
    Model        & A2             & B1             & B2             & C1             & Macro \\
    \midrule
    CASA         & 0.553          & 0.351          & \textbf{0.290} & \textbf{0.554} & \textbf{0.437} \\
    CASA-Crisper & \textbf{0.485} & \textbf{0.335} & 0.322          & 0.617          & 0.440 \\
    \bottomrule
  \end{tabular}
\end{table}

CASA achieves per-part RMSEs of 0.476, 0.454, 0.490, and 0.444 for P1, P3, P4, and P5, respectively. Performance is weaker on P1 and P4, consistent with Lin et al. \cite{linSessionLevelSpokenLanguage2026}. A qualitative analysis of the ASR transcripts, conducted with a co-author in applied linguistics, suggests that task format may partly explain this pattern: short personal questions in P1 and process descriptions in P4 provide less freedom to demonstrate broad linguistic and content proficiency than the opinion-based tasks in P3 and P5. Analyzing the same task formats, Banno et al. \cite{banno_NLP_ASA2025} similarly found that P1 and P4 place greater emphasis on fluency, whereas P3 and P5 place greater weight on thematic development and other content-related dimensions. P1 and P4 may therefore provide fewer discriminative textual cues. In CASA, the task embeddings in the Transformer block proved ineffective: they remained close to their zero initialization, contributing little task information to the acoustic representation. Future work could investigate separate prediction heads or graders for more constrained and open-ended tasks.

To assess the stability of the SOTA-level result, we performed 9 additional runs for each configuration in Table~\ref{tab:variance}: 4 with seed 1011 and 5 with distinct seeds. Here, 4e-4 doubles the acoustic-branch LoRA learning rate, while aux-0 sets the auxiliary-loss weight to 0. The auxiliary head provides a small but consistent improvement across repeated runs, reducing mean RMSE by 0.004 and helping CASA reach SOTA-level performance. In two additional single-run ablations, blocking either the auxiliary-loss gradient or the soft-token-path (main-loss) gradient from reaching the acoustic encoder degraded performance, suggesting the two paths provide complementary supervision to the shared acoustic encoder. Doubling the learning rate to 4e-4 unexpectedly yielded an RMSE of 0.350 on the first run, but subsequent runs did not reproduce it, indicating an unstable configuration rather than a better one.

\begin{table}[t]
  \caption{RMSE variability on the S\&I test set over 10 runs. The 95\% CI is calculated for the mean.}
  \label{tab:variance}
  \centering
  \begin{tabular}{lcccc}
    \toprule
    Model & Mean  & Median & Min--Max          & 95\% CI \\
    \midrule
    CASA  & 0.363 & 0.362  & 0.357--0.377   & [0.359,\,0.367] \\
    aux-0  & 0.367 & 0.366  & 0.362--0.376  & [0.364,\,0.370] \\
    4e-4  & 0.378 & 0.376  & 0.350--0.402   & [0.364,\,0.392] \\
    \bottomrule
  \end{tabular}
\end{table}

For CASA, even runs using the same seed produced test RMSEs ranging from 0.357 to 0.365, indicating nondeterminism in the training pipeline. Most distinct seeds produced comparable results, except for the poorer run with seed 6066. The 0.358 result in Table~\ref{tab:results} was obtained from the first run of the configuration and was selected on the development set, rather than from the best repeated run. Full run-level results are available in our repository.

Although efficiency remains a primary goal, we argue that LLMs offer substantial value for ASA beyond their parameter scale. In particular, they can adapt to different assessment tasks~\cite{maAssessmentL2Oral2025} and provide speech content assessment and feedback without task-specific training~\cite{phan24_interspeech}. As an illustration, we use CASA's Qwen3.5-2B for few-shot content validation, prompting it at inference time to judge whether an answer addresses the given question. Replacing the original questions with an unrelated question about nuclear reactors, the LLM flags 99.9\% of responses as not addressing the question; pairing responses with real questions from different parts yields 97.3\%. This is also practical: judgments took just under 0.1\,s per response. Such validation requires reasoning over the semantic relationship between question and answer, which is not available to an acoustic-only grader~\cite{phanOneWhisperGrade2025}. We noted that the same judge could also be prompted to flag A2-level content even from ASR transcripts, though inconsistently; we leave this as a direction for future work.

\section{Conclusion}
\label{sec:conclusion}

In this work, we introduced CASA, a model architecture that explicitly separates speech delivery from speech content in ASA. CASA achieves SOTA-level performance on the S\&I test set while using approximately half the inference parameters of comparable LLM-based systems. Our analyses demonstrate the complementary roles of the acoustic and content branches, identify task-specific modeling and verbatim transcription as promising directions for further improvement, and show that the LLM branch can also support training-free content validation. By releasing our efficient, general-purpose architecture and experimental results, we aim to support reproducible research and further investigation of acoustic and linguistic information in ASA.

\section{Acknowledgements}
We would like to thank the following projects and funding agencies: Research Council of Finland through the funding to ``DigiTala in action - Automatic speaking assessment supporting integration'' (grant no 365232, 365233), and ``Automatic assessment of spoken interaction in second language'' (grant no 355586, 355587). The computational resources were provided by Aalto ScienceIT.

\bibliographystyle{IEEEbib}
\bibliography{myRefs}

\end{document}